# INDEXITY: a web-based collaborative tool for medical video annotation

Jean-Paul Mazellier [a,b], Méline Bour-Lang [a], Sabrina Bourouis [a], Johan Moreau [c], Aimable Muzuri [d], Olivier Schweitzer [c], Aslan Vatsaev [a], Julien Waechter [a], Emilie Wernert [a], Frederic Woelffel [a], Alexandre Hostettler [c], Nicolas Padoy [a,b], Flavien Bridault [c]

a . IHU Strasbourg, France

b . ICube, University of Strasbourg, CNRS, France

c . IRCAD France, Strasbourg, France

d . IRCAD Africa, Strasbourg, Rwanda

## 0 ) Abstract

This technical report presents Indexity 1.4.0, a web-based tool designed for medical video annotation in surgical data science projects. We describe the main features available for the management of videos, annotations, ontology and users, as well as the global software architecture.

## 1 ) Introduction

Surgical Data Science (SDS) [Maier-Hein2022] is a research field applying general Data Science methods to the medical field. For many Data Science tasks, high quality and large dataset production is necessary to train algorithms. This is particularly true in the Deep Learning and Big Data era. But if the creation of such datasets is already a gigantic effort for general purpose tasks (object detection, human pose estimation, etc.), the annotation of surgical datasets is also complicated by the highly sensitive nature of the involved data and by the need for clinical expertise, which is especially rare and costly. General annotation tools are generally either dedicated to data scientists or non-expert annotators. Their access can involve the installation of the program on a local computer and in some cases the compilation of the source code. Also, the user interface is not necessarily optimized for non-informatician users. On top of that, data sharing (medical material and annotations) can be complicated and would need particular server setup and access. These points are often a deal breaker outside of the data scientist community.

Many annotation tools, either proprietary [V7, Superannotate] or open-sourced [CVAT, LabelMe], are emerging in order to propose a simpler access to non-informatician communities. Nevertheless, these annotation tools are quite general and lack features addressing the SDS particular needs such as collaborative annotation features, study group features, medical ontology implementation, etc.

In this context, we present a web-based annotation tool denoted as « Indexity » which was designed to be user-friendly for the medical community. In this technical report, we detail the general structure and features of this tool. Indexity has been co-developed by IHU Strasbourg and IRCAD during the period of April 2017 (version 0.0.1) to January 2021 (version 1.4.0). The source code is under BSD3 licensing and public repository is available at https://github.com/CAMMA-public/indexity.

# 2 ) Technical description of Indexity 1.4.0 annotation tool

### A ) Features

#### i . Landing page and Signup/Login views

In order to ensure data safety, connection to Indexity 1.4.0 is protected by a username/password combination. A manual activation is performed by administrators to ensure user information.

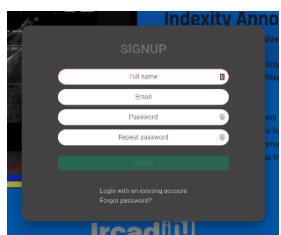
(a) Signup

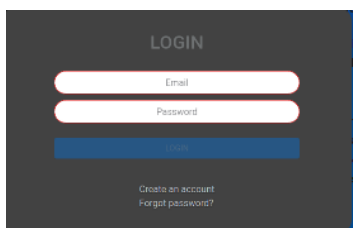
(b) Login

*Figure 1. (a) Signup ; and (b) login interfaces*

#### ii . Video list view

With Indexity, we focus on video annotation. A video owner can upload (if granted permission) a video on the Indexity platform. All videos are then available under a thumbnailed patched view (Global Video View) as depicted in Figure 2. Different types of information, such as video annotation status, can be retrieved in this view, and operations such as filtering (bookmarked videos), searching, rename/delete, etc, can be performed.

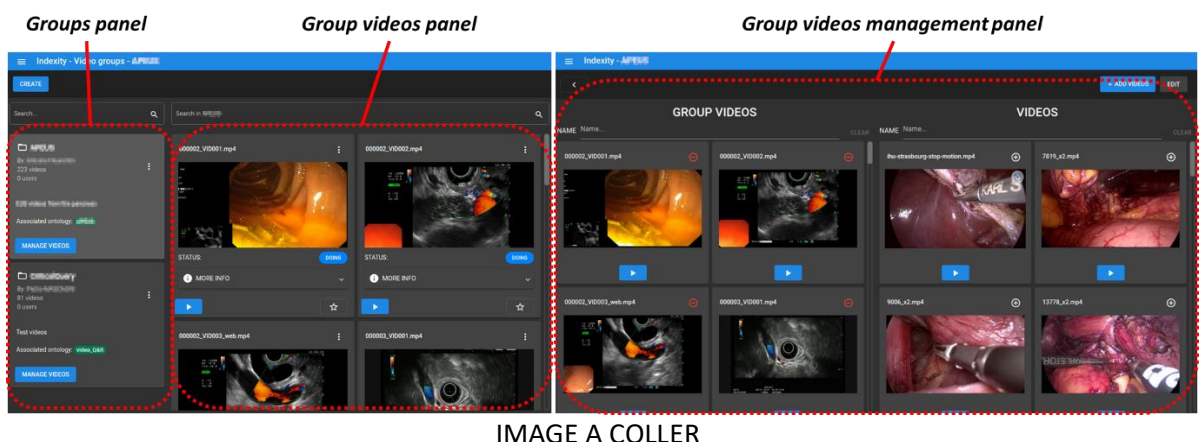

IMAGE A COLLER

*Figure 2. Global videos view. Once a user (with corresponding permission) uploads a video, it is automatically available in this workspace where different functionalities/types of information are available.*

#### iii. Ontology management

A global Indexity ontology list is available. It offers the possibility to define different types of labels:

- Temporal: **Phase / Action / Event**. The different types are designed to differentiate the possible types of annotation of temporal information.

- Spatio-temporal: **Structure**. This offers the possibility to draw bounding boxes around an object instance over time during a user-defined period.

For all the label types, users can define the label properties « Name » and « Color ».

**iv. Annotation view**

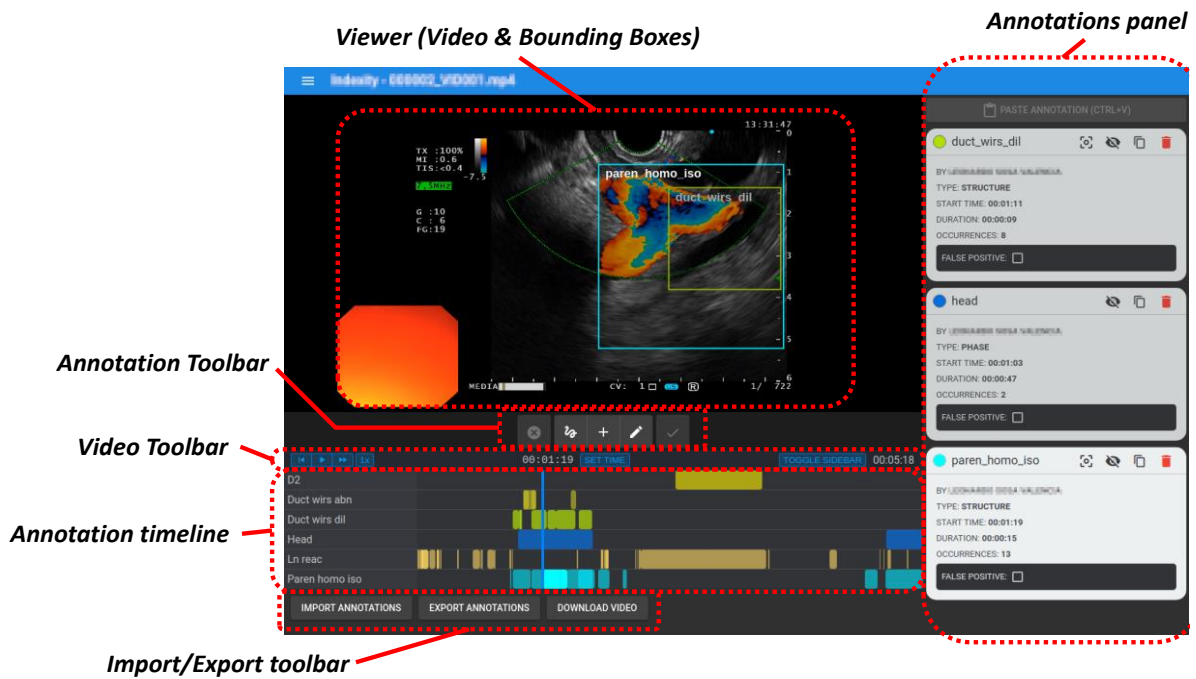

*Figure 3. Annotation view in Indexity 1.4.0. In addition to the video viewer, a video toolbar allows for video navigation. Correspondingly an annotation timeline offers a synthetic view of the (spatio-)temporal annotation set. On the right, active annotations instances (depending on time cursor position) are available for different actions (copy, paste, hide, delete, …). Annotations (in json format) can be exported/imported with a dedicated button at the bottom. An annotation toolbar allows to select the two main types of annotations (temporal or spatio-temporal).*

Inspired by the audiovisual approach, the temporal definition of the annotations is based on temporal information with a millisecond precision. All the annotations are shown on the timeline panel with the possibility to zoom in/out (with scroll bars for navigation if needed, horizontally for time navigation or vertically for labels navigation). To offer a larger timeline view, a « toggle sidebar » button allows to hide the labels name column.

An (spatio-)temporal annotation can be created by selecting the corresponding tool in the annotation toolbar:

- Temporal annotation: By clicking on the temporal annotation button and after selecting/creating a label, a start and stop time must be defined. The start time is automatically associated with the current time cursor position at annotation creation, the stop time must be validated by moving the time cursor (either by clicking on the timeline or by playing/pausing the video with video toolbar buttons).
- Spatiotemporal annotation: A spatiotemporal annotation can be created in the very same way. At start time, the first bounding box (BB) must be drawn and can then be updated through the temporal duration of the annotation. The updates are recorded by moving the BB while playing the video. Each time the BB properties (position and/or dimensions) change, a temporal key point is created. Linear interpolation of position and dimensions can be operated in-between two key points. Another option available in creating a spatiotemporal annotation is to use the Kernelized Correlation Filter (KCF) tracking tool from the OpenCV library [OpenCV]: by initiating the object to track at the beginning of the annotation (by drawing a BB to this object), the tracking BB can be automatically proposed this way.

Every annotation has some properties that can be retrieved in the « Annotations panel ». These properties can be the occurrence number (sequencing number of a particular label appears in continuous section of the video over time), false positive (user defined for collaborative cross-check), start & duration, identifier of the user who has created the annotation. Spatiotemporal annotations also offer the possibility to have a written label displayed on the video viewer (in addition to BB display) as well as the activation of an automated tracker as mentioned earlier.

After creation, annotations can be further edited. Editing includes label modification. For spatiotemporal annotation, it is also possible to temporally divide an annotation in two annotations as well as to correct the BB position and/or dimensions while keeping the initial timespan.

As Indexity is intended to be collaborative, websockets are used to display in real-time annotations to any user. So, if a user is creating an annotation on a particular video, this annotation is directly visible to any user working on the same video. It avoids issues when multiple users are working on the same material.

### v. Administration and Roles

In Indexity 1.4.0, three roles are defined in order to grant particular permission sets to users. The admin role (allowing all permissions), moderator (allowing video management and annotation), annotator (allowing only annotation). These roles are depicted in Figure 4.

| | admin | moderator | annotator |
|---|---|---|---|
| annotate video | ✓ | ✓ | ✓ |
| add video | ✓ | ✓ | |
| delete video | ✓ | ✓ | |
| add user | ✓ | | |
| delete user | ✓ | | |

*Figure 4. Role definition in Indexity 1.4.0*

### vi. Group view

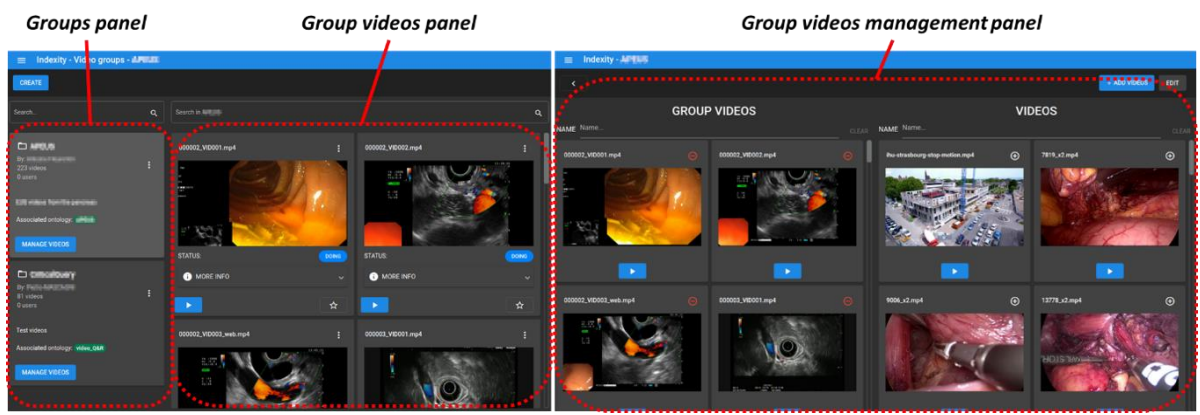

*Figure 5. Group view (left) and Group videos management view (right).*

All videos uploaded in Indexity 1.4.0 are available to any user. In order to simplify the structuration of the collaborative annotation work, groups can be created. They basically act as folders where subsets of available videos can be pointed to. Also, a subset of ontology can be associated to the group (in order to restrict the available ontology only to the particular task the group is intended for). Note that a video can be pointed to from different groups at the same time, avoiding a duplication of the video in central storage. On the contrary, annotation sets are linked to a particular group and are not shared in between groups.

### B ) Architecture

In Figure 6 is presented the general overview of Indexity 1.4.0 architecture. It is mainly composed of (a) a back-end for managing requests, file storage and annotation database; (b) a front-end the web-browser interface; and (c) a restfull-API.

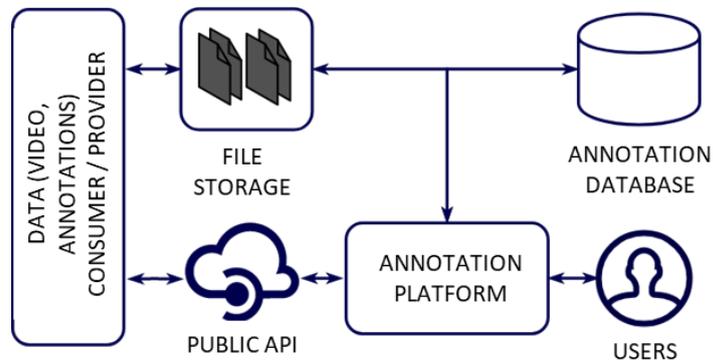

*Figure 6. Indexity v1.4.0 general architecture*

Indexity 1.4.0 is composed of several services:

- **Indexity-ui**: it is the web application of Indexity. It allows to display the application browser for annotators and handles interactions with indexity-api.

- **Indexity-api**: it is the Indexity backend. It manages the authentication system and users as well as the storage of videos and annotations. It also provides a REST API.

- **Indexity-admin**: it is the administration application which is displayed in the browser of the application for administrators and allows user management.

- **Indexity-structure-tracker**: it is an automatic tracking service aimed at providing an automated way of tracking spatial annotations based on KCF algorithm [OPENCV]. It also communicates with the API.

- **PyIndexity**: it is a Python module for simplifying the use of the Indexity API. It facilitates the management of videos and annotations.

### C ) Dependencies

Indexity 1.4.0 is based on open-source frameworks. Hereafter are presented the main framework dependencies of the different Indexity services. Docker and Kubernetes are used to deploy the following services:

| Services \ Frameworks | Angular | NgRx | NestJS | Python | OpenCV |
|---|---|---|---|---|---|
| Indexity-ui | V9.0 | V9.2 | - | - | - |
| Indexity-api | - | - | V6.10 | - | - |
| Indexity-admin | V8.1 | - | - | - | - |
| Indexity-structure-tracker | - | - | V6.10 | V3.6 | V4.2 |
| PyIndexity | - | - | - | V3.7 | - |

For database management, Postgres (v11) and Redis (v5) are used. The detailed architecture of theservices and databases is reported in Figure 7.

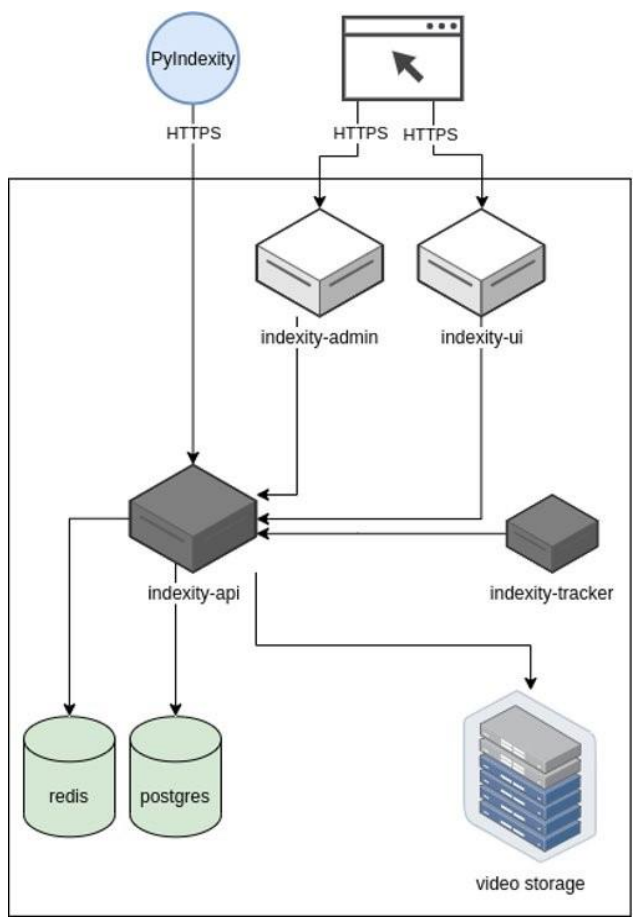

Figure 7. Indexity (v1.4.0) detailed architecture

## 3 ) Conclusion

To address the lack of a convenient annotation tool for medical videos, Indexity has been developed as a web-based service allowing collaborative dataset generation. A initial set of functionalities have been implemented in Indexity 1.4.0 in order to provide clinicians and data scientists with a tool to produce high-quality annotations.

# 4 ) Acknowledgment

This work was supported by French state funds managed within the "Plan Investissements d'Avenir" and by the ANR (reference ANR-10-IAHU-02) as well as by Banque Publique d'Investissement through the « CONDOR » project.